\pdfoutput=1

\documentclass[11pt]{article}
\usepackage[preprint]{acl}

\usepackage{times}
\usepackage{latexsym}
\usepackage{graphicx} 
\usepackage{amsmath}
\usepackage{mathrsfs}
\usepackage{booktabs}
\usepackage{multirow}
\usepackage{cleveref}
\usepackage{array}
\usepackage[T1]{fontenc}

\usepackage[utf8]{inputenc}

\usepackage{microtype}

\usepackage{inconsolata}

\title{Tree of Reviews: A Tree-based Dynamic Iterative Retrieval Framework for Multi-hop Question Answering}

\author{
  \textbf{Jiapeng Li\textsuperscript{1}},
  \textbf{Runze Liu\textsuperscript{2}},
  \textbf{Yabo Li\textsuperscript{1}},
  \textbf{Tong Zhou\textsuperscript{1}},
  \textbf{Mingming Li\textsuperscript{1}},
  \textbf{Xiang Chen\textsuperscript{1}}
\\
  \textsuperscript{1}Tencent Inc., Shenzhen, China \\
  \textsuperscript{2}Harbin Institute of Technology, Harbin, China
\\
    \{montereyli, kerli, tongzhou, merakili, joshuaxchen\}@tencent.com, \\
    rzliu@ir.hit.edu.cn
}

\begin{document}
\maketitle
\begin{abstract}
Multi-hop question answering is a knowledge-intensive complex problem. Large Language Models (LLMs) use their Chain of Thoughts (CoT) capability to reason complex problems step by step, and retrieval-augmentation can effectively alleviate factual errors caused by outdated and unknown knowledge in LLMs. Recent works have introduced retrieval-augmentation in the CoT reasoning to solve multi-hop question answering. However, these chain methods have the following problems: 1) Retrieved irrelevant paragraphs may mislead the reasoning; 2) An error in the chain structure may lead to a cascade of errors.

In this paper, we propose a dynamic retrieval framework called \textsc{Tree of Reviews (ToR)}, where the root node is the question, and the other nodes are paragraphs from retrieval, extending different reasoning paths from the root node to other nodes. Our framework dynamically decides to initiate a new search, reject, or accept based on the paragraphs on the reasoning paths. Compared to related work, we introduce a tree structure to handle each retrieved paragraph separately, alleviating the misleading effect of irrelevant paragraphs on the reasoning path; the diversity of reasoning path extension reduces the impact of a single reasoning error on the whole. We conducted experiments on three different multi-hop question answering datasets. The results show that compared to the baseline methods, \textsc{ToR} achieves state-of-the-art performance in both retrieval and response generation. In addition, we propose two tree-based search optimization strategies, pruning and effective expansion, to reduce time overhead and increase the diversity of path extension. We will release our code.


\end{abstract}

\section{Introduction}

\begin{figure}[ht]
  \centering
  \includegraphics[width=0.5\textwidth]{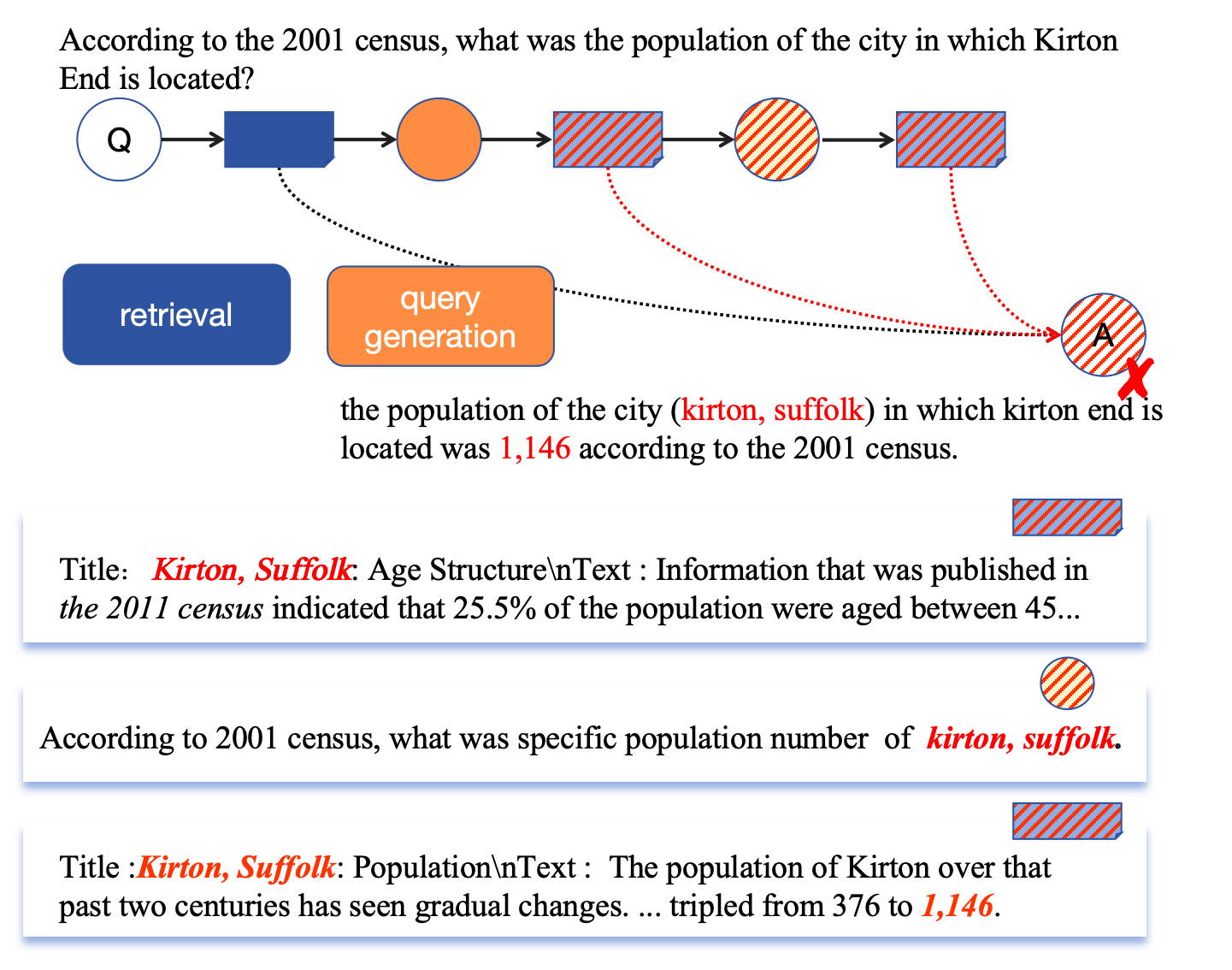}
  \caption{The chain-like iterative retrieval process faces the issue of error accumulation. The example shows how irrelevant retrieval results affect subsequent retrievals and the generation of new queries, ultimately leading to incorrect answers.}
  \label{fig:introduction}
\end{figure}

Large Language Models (LLMs) have demonstrated the capacity for multi-step reasoning \citep{NEURIPS2022_9d560961}. This is achieved by generating intermediate reasoning steps, a process known as the chain of thoughts (CoT) \citep{NEURIPS2022_8bb0d291}. However, despite their advanced reasoning capabilities, LLMs are sometimes prone to generating incorrect reasoning steps. These inaccuracies can be attributed to the lack of current knowledge within their parameters or the erroneous retrieval of information encoded in their weights \citep{maynez-etal-2020-faithfulness}. In response to this issue, arguments for LLMs with knowledge from external data sources have emerged as a promising approach, attracting increased attention from researchers \citep{DBLP:journals/corr/abs-2301-12652,jiang-etal-2023-active,trivedi-etal-2023-interleaving}.


In some typical question-answering tasks, retrieval-augmented language models utilize a one-time retrieval method \citep{JMLR:v24:23-0037, NEURIPS2020_6b493230}. However, these methods are not satisfied for multi-hop questions, necessitating a more nuanced approach to acquiring comprehensive knowledge. Such questions often involve indirect facts that may exhibit minimal lexical or semantic correlation with the question but are essential for reaching accurate answers. For instance, to answer the question,\texttt{‘According to the 2001 census, what was the population of the city in which Kirton End is located?’}. First,  we need to know that \texttt{'Kirton End is located in Boston'}, then look up \texttt{'the population of Boston according to the 2001 census'}. This process highlights the necessity of iterative retrieval, underscoring the limitations of one-time retrieval strategies in addressing complex informational needs.

Iterative retrieval involves conducting multiple turns of retrieval, each guided by newly generated sub-questions \citep{press-etal-2023-measuring}, the most recent response \citep{shao-etal-2023-enhancing}, or an intermediate reasoning step \citep{trivedi-etal-2023-interleaving}. As illustrated in Figure \ref{fig:introduction}, these methods employ a sequential, chain-like process alternating between retrieval and query generation. While these approaches demonstrate superior performance compared to one-time retrieval, the chain-like nature of the process is susceptible to cascading errors. A local error at any step, whether due to inappropriate retrieval or query generation, can affect subsequent steps, culminating in incorrect responses. Such errors underscore the inherent vulnerability of iterative retrieval methods, highlighting a critical challenge in achieving reliable knowledge extraction.


To this end, drawing inspiration from previous work that leveraged a tree-like reasoning process to enhance reasoning capability \citep{DBLP:journals/corr/abs-2305-10601}, this paper introduces \textsc{Tree of Reviews(ToR)} a dynamic, tree-based iterative retrieval framework. We follow the \texttt{retrieve-and-read} paradigm \citep{DBLP:journals/corr/abs-2101-00774}, the retriever initially retrieves knowledge for the question, and the reader utilizes this retrieved knowledge to generate a response.

In detail, we construct a tree with the initial question serving as the root and individual paragraphs as other nodes during the retrieval phase. Each node contains a single paragraph, mitigating the risk of diverging the reasoning process due to irrelevant information. A paragraph review block within this structure evaluates each node to determine the subsequent action—further retrieval, acceptance, or rejection. Each accepted path is referred to as a piece of evidence. We propose three evidence fusion methods, allowing the reader to utilize evidence from various paths to generate the final response. Incorporating a tree structure into the retrieval process has enhanced the performance of paragraph retrieval and answer accuracy.

To further enhance the search efficiency of ToR, we advocate for two tree-based search optimization strategies: \textit{pruning}, which aims to diminish the frequency of unproductive search initiations, and  \textit{effective expansion}, designed to refine query generation for improved retrieval paragraphs.

Experiments on three different multi-hop question answering datasets show that our proposed method achieves state-of-the-art performance in both retrieval and response generation. 
 
Our contributions include:
\begin{itemize}
    \item We propose a dynamic retrieval framework named \textsc{Tree of Reviews (ToR)}, which integrates a tree structure into the iterative retrieval process. This method mitigates the negative impact associated with the inherent vulnerabilities of chain-like retrieval methods.
    \item We propose two tree-based search optimization strategies: pruning and effective expansion. These strategies demonstrate significant improvements in both retrieval quality and efficiency. These efforts offer valuable insights for the optimization of iterative retrieval methods.
    \item Our method achieves state-of-the-art performance in both retrieval and response generation on three different multi-hop question answering datasets. Extensive experiments have conclusively demonstrated the effectiveness of our method.
\end{itemize}

\section{Related Work}
\subsection*{Supervised Multi-hop Question Answering}
Some researchers have investigated iterative retrieval for multi-hop question answering in fully supervised settings. \citet{DBLP:conf/iclr/DasDZM19} generate a new query representation by utilizing the current query and the current state of the reader, and initiate iterative retrieval. \citet{feldman-el-yaniv-2019-multi} adopt a similar approach, in which a fusion module is designed in the new query generation stage to ensure sufficient interaction. \citet{qi-etal-2019-answering} employ a supervised generator to generate new queries based on the query and historical passages, and iteratively conduct retrieval. \citet{DBLP:journals/corr/abs-2112-09332} utilize GPT3 to answer long-form questions by simulating human browsing behavior. Although supervised models perform well on multi-hop datasets like HotpotQA, they rely on expensive manual annotation and specific training. However, in practical application scenarios like New Bing and PerPlexity.AI, the indexed document scope is broader, and the retrieval source is updated in real-time. In this case, the supervised models are likely to fail.


\subsection*{Retrieval-Augmentation for Complex Problems} 
The Retrieval-Augmented Generation (RAG) system typically retrieves additional knowledge from specific corpora, such as Wikipedia, to alleviate the hallucination problem of Large Language Models (LLMs), thereby significantly enhancing the performance of LLMs in various tasks \citep{NEURIPS2020_6b493230,pmlr-v119-guu20a,ram-etal-2023-context}. Early research on RAG typically employs a one-step retrieval approach, which is ineffective in addressing composite problems. To tackle composite problems, Self-Ask \citep{press-etal-2023-measuring} poses sub-questions before answering the main question, optimizing complex composite problems through multiple retrievals.IRCoT \citep{trivedi-etal-2023-interleaving} triggers retrieval on each sentence of the CoT. ITER-RETGEN \citep{shao-etal-2023-enhancing} connects the complete CoT reasoning steps generated in the previous turn with the original question for the next turn's generation query. However, these methods all adopt a chain-like structure for reasoning. If an error occurs at any step in the reasoning path, it could potentially cause the reasoning path to deviate.

\subsection*{Tree-like Reasoning for Complex Problems}
The tree is an efficient structure for solving complex reasoning problems \citep{DBLP:journals/corr/abs-2305-10601}. Tree of Thought(ToT) enhances the problem-solving capabilities of Large Language Models (LLMs) by introducing a tree-like structure during the reasoning process, simulating the human problem-solving process. This allows the model to consider multiple reasoning paths and self-evaluate to decide the following action. \citet{DBLP:conf/iclr/AsaiHHSX20} trained a retriever that dynamically retrieves information from Wikipedia graphs. However, this method relied on a hyperlink graph constructed from Wikipedia, which fails when the path related to the problem is not included. Some researchers decompose complex problems into a static problem tree with several sub-problems. Then, answer each sub-problem by utilizing language models and additional retrieval information \citep{cao-etal-2023-probabilistic} or calculating the probability of reasoning paths  \citep{zhang-etal-2023-reasoning}, ultimately solving the complex problem. However, the decomposition of the question and the construction of the tree lack the assistance of external knowledge and information on the reasoning path, which can easily lead to incorrect decomposition, possibly affecting the correctness of the final answer.

In contrast, our work is the first to propose a retrieval framework that uses a tree-like structure to dynamically initiate requests based on external knowledge and information on the reasoning path. LLMs can decide dynamically whether to initiate further retrieval and what requests to generate based on this information. We have designed two search optimization strategies to reduce the time overhead of tree structure searching and enhance the diversity of initiating requests: pruning and effective expansion.


\begin{figure*}[thbp]
  \centering
  \includegraphics[width=\textwidth]{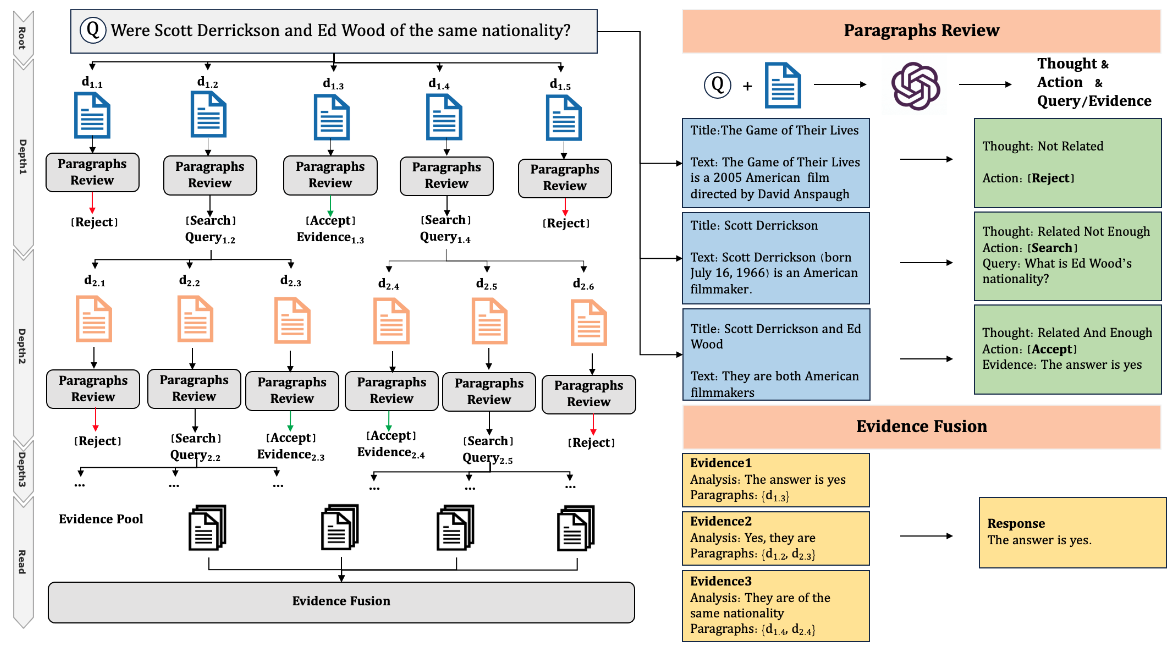}
  \caption{The left side illustrates the overall \textsc{ToR} framework (introduced in Section \ref{chapter3-1}). The upper right half illustrates \textbf{Paragraphs Review} a fundamental component of \textsc{ToR} (introduced in Section \ref{chapter3-2}). The lower right half illustrates  \textbf{Evidence Fusion} a method for more effectively utilizing retrieved information for reading (introduced in Section \ref{chapter3-3}). }
  \label{fig:TOR Framework}
\end{figure*}

\section{Tree of Reviews Framework}
\subsection{Overall}
\label{chapter3-1}
The task is to answer a multi-hop question \textit{Q} based on a retrieval corpus \textit{D}. As illustrated in Figure \ref{fig:TOR Framework}, we introduce \textsc{Tree of Reviews(ToR)}, a tree-based dynamic retrieval framework. In this framework, the root node is the question \textit{Q}, while each subsequent node is a paragraph with a paragraphs review block (Sec.\ref{chapter3-2}). These blocks dynamically judge whether to stop or continue the search based on all paragraphs along the path from the root node to the current node. If there are enough paragraphs to answer \textit{Q}, the model will use them to produce evidence and add it to the evidence pool. Upon reviewing all paths, the reader generates the final response to \textit{Q} based on the evidence in the evidence pool. We propose three evidence fusion strategies (Sec.\ref{chapter3-3}) to fully use information from diverse reasoning paths. Additionally, we propose two tree-based search optimization methods (Sec.\ref{chapter3-4}): pruning and effective expansion. These methods aim to enhance the search efficiency of the \textsc{ToR} framework.

\subsection{Paragraphs Review }\label{chapter3-2}
The \textsc{ToR} framework initiates multiple retrievals, utilizing both the original question \textit{Q} and new queries generated by the paragraphs review block.  We use a dense retriever proposed by \citet{JMLR:v24:23-0037}  to encode both the query and the retrieval corpus \textit{D} and then compute the similarity score of their embeddings by cosine similarity. 

We expand a child node for each retrieved paragraph individually. The paragraphs review block selects an action based on the question \textit{Q} and paragraphs along the path. The block is designed to execute the following steps:\\
(i) Judging whether the paragraphs are relevant to the question \textit{Q}.  Relevance means that paragraphs contain partial information to answer the question.\\
(ii) Judging whether the paragraphs have enough information, which means that paragraphs contain all information to answer the question\\
(iii) If it is not relevant, then action selection is \texttt{[Reject]} then stop search; if it is relevant but not enough, action selection is \texttt{[Search]} then generate a new query and retrieval with it; if it is relevant and enough, then action selection is \texttt{[Accept]} then stop search. Once LLM accepts a reasoning path, it will be asked to answer the original question Q based on all the documents on this reasoning path. This answer is called a \textbf{brief analysis}. The accepted reasoning path combined with the brief analysis form a piece of \textbf{evidence}. \\

Utilizing the method above, we iteratively execute retrieval and review through the depth-first search until each reasoning path is either accepted, rejected, or reaches its maximum search depth.
\subsection{Evidence Fusion}\label{chapter3-3}
The evidence pool contains some pieces of evidence. The QA reader will generate a response according to the evidence pool. We propose three simple methods for evidence fusion:\\
\textbf{Analysis-based Fusion}: The reader generates a response only according to the brief analysis.\\
\textbf{Paragraph-Based Fusion}: The reader generates a response only according to the paragraphs.\\
\textbf{Evidence-based fusion}: The reader generates a response according to both of them. \\
See Table \ref{tab:P-based-EF} to \ref{tab:E-based-EF} for details.

\begin{table*}[thbp]
  \centering
\begin{tabular}{ccccccc}
\toprule
\multirow{2}{*}{Method} & \multicolumn{3}{c}{GPT-3.5-Turbo} & \multicolumn{3}{c}{GPT-4-Turbo}\\
\cmidrule(lr){2-4} \cmidrule(lr){5-7}
 &  \multicolumn{1}{c}{HotpotQA} &  \multicolumn{1}{c}{2WikiMQA} & \multicolumn{1}{c}{MuSiQue} & \multicolumn{1}{c}{HotpotQA} & \multicolumn{1}{c}{2WikiMQA} & \multicolumn{1}{c}{MuSiQue} \\
\midrule 
OneR & 44.3 & 45.8 & 23.2 & 44.3 &  45.8 & 23.2\\
ReAct & 44.6 & 48.0 & 25.2 & 51.3 &  46.1 & 35.5\\
Self-AsK & 44.0 & 50.7 & 25.9 & 52.9 &  59.5 & 37.2\\
ITER-RETGEN & \underline{50.6} & \underline{51.1} & \underline{27.2} & \underline{60.5} &  \underline{67.4} & \underline{47.0}\\
IRCoT & 46.0 & 46.5 & 25.2 & 53.3 & 53.9 & 36.5\\
\midrule 
CoR & 47.9 & 47.6 & 25.8 & 61.0 &  62.4 & 39.4\\
ToR & \textbf{53.1} & \textbf{51.8} & \textbf{29.5} & \textbf{73.8} &  \textbf{79.4}& \textbf{48.5}\\
\bottomrule
\end{tabular}
\caption{ Paragraphs recall@15 on multi-hop question answering datasets. We highlight the best results in bold and underline the best results among other methods.}
\label{tab:retrieval result}
\end{table*}

\subsection{Tree-Based Search Optimization}\label{chapter3-4}
Although the tree structure can explore more diverse reasoning paths and reduce failures caused by a single reasoning path, it introduces longer time overheads. Therefore, we propose pruning and effective expansion to reduce redundancy and irrelevant expansion in the search process while guaranteeing expansion diversity.

\textbf{Pruning} aims to reduce the initiation of invalid searches. We propose two methods: \textbf{Relevance Pruning} and \textbf{Repetitive Pruning}. Relevance pruning is conducted at the paragraphs review block, where the model judges whether the paragraphs are relevant to the question and subsequently prunes the paths expanded from irrelevant paragraphs. Repetitive pruning is conducted after retrieval, where it matches the paragraph ID of retrieved paragraphs and received paragraphs in the evidence pool. If any retrieved paragraph is already in the evidence pool, it is pruned.

\textbf{Effective expansion} aims to optimize the effectiveness and diversity of paragraphs review block initiating new queries. We adopt \textbf{CoT Expansion} and \textbf{Missing Paragraph Completion Expansion (MPC)}. CoT expansion allows the model to think step by step, identify missing information in current paragraphs and generate a new query based on this missing information. MPC expansion enables the model to complete the missing information in paragraphs using its internal knowledge and to use the newly generated paragraph as a new query. See Table \ref{tab:PR-CoT} and Table \ref{tab:MPC} for details.

\section{Experiments}

\subsection{Datasets}
We conducted experiments on three multi-hop reasoning datasets: HotpotQA with Fullwiki setting \citep{yang-etal-2018-hotpotqa}, 2Wiki-MultiHopQA \citep{ho-etal-2020-constructing}, and the answerable subset of MuSiQue \citep{trivedi-etal-2022-musique}. For HotpotQA and 2Wiki-MultiHopQA, we used the Wikipedia dump from December 2018 as the retrieval source, while for MuSiQue, we used the Wikipedia dump from December 2021. Following the work of previous researchers \citep{shao-etal-2023-enhancing}, we used the first 500 questions from the development sets of these datasets for retrieval and response generation performance evaluation in Table \ref{tab:retrieval result} and Table \ref{tab:answer result}. Then, randomly selected 100 questions from the remaining part for hyperparameter tuning in Table \ref{tab:evidence fusion} to Table \ref{tab:hyperparameters}.

\subsection{Evaluate Setting} \label{chapter 4-2}
We evaluated \textsc{ToR} from retrieval quality and generation quality. For the retrieval metric, we followed \citet{trivedi-etal-2023-interleaving}, allowing different retrieval systems to return up to 15 paragraphs and calculating the recall of golden paragraphs; this is referred to as recall@15\footnote{In this work if the number of retrieved paragraphs exceeds 15, we re-rank the evidence in the evidence pool based on the similarity between the evidence and the final response. We select the top 15 paragraphs according to their similarity scores.}. We used exact match (EM) and F1 score for the generation metric.

\subsection{Baselines}
Given the disparity in retriever, reader, and test samples used by the baseline methods, a fair comparison becomes challenging. Therefore, we followed \citet{shao-etal-2023-enhancing} used \textbf{Contriver} \citep{JMLR:v24:23-0037} as our retriever. \textbf{GPT-3.5-Turbo}(gpt-3.5-turbo-0125) and \textbf{GPT-4-Turbo}(gpt-4-1106-preview) \citep{NEURIPS2022_b1efde53, openai2023gpt4}  were used as the base models to implement the following baseline methods. The format of prompts and few-shot settings are adopted as presented in their papers. We retrieved top-5 paragraphs for each query, and for baselines involving multi-turn iterations, we set the maximum number of turns to 3. \\
\begin{table*}[thbp]
  \centering
\begin{tabular}{ccccccccccccc}
\toprule
\multirow{3}{*}{Method} & \multicolumn{6}{c}{GPT-3.5-Turbo} & \multicolumn{6}{c}{GPT-4-Turbo}\\
\cmidrule(lr){2-7} \cmidrule(lr){8-13}
 &  \multicolumn{2}{c}{HotpotQA} &  \multicolumn{2}{c}{2WikiMQA} & \multicolumn{2}{c}{MuSiQue} & \multicolumn{2}{c}{HotpotQA} & \multicolumn{2}{c}{2WikiMQA} & \multicolumn{2}{c}{MuSiQue} \\
 \cmidrule(lr){2-3} \cmidrule(lr){4-5} \cmidrule(lr){6-7} \cmidrule(lr){8-9} \cmidrule(lr){10-11} \cmidrule(lr){12-13}
  &  EM & F1 &  \multicolumn{1}{c}{EM} & \multicolumn{1}{c}{F1} & EM & F1 & EM & F1 & EM & F1 & EM & F1  \\
\midrule
\multicolumn{13}{c}{Without Retrieval} \\
\midrule
Direct & 28.2 & 37.7 & \underline{27.6} & 31.8 &  9.6 & 18.2 & 40.6 & 52.2 &  38.4 & 47.6 & 25.9 & 36.0 \\
CoT & 27.8 & 38.8 & 26.7 & \underline{33.6} &  \underline{\textbf{17.2}} & \underline{\textbf{25.1}} & 39.6 & 53.3 &  \underline{42.2} & 51.8 & 24.0 & 36.8 \\
\midrule
\multicolumn{13}{c}{With Retrieval} \\
\midrule 
OneR-Direct & 25.0 & 33.4 & 20.6 & 23.8 &  5.0 & 10.3 & 40.2 & 53.6 &  32.6 & 42.4 & 26.2 & 37.4 \\
OneR-CoT & 24.8 & 32.1 & 14.0 & 18.7 &  5.0 & 11.2 & 39.6 & 52.3 &  36.6 & 47.0 & 22.8 & 34.9 \\
ReAct & 25.8 & 37.2 & 14.6 & 24.3 &  2.2 & 7.7 & 34.8 & 47.5 &  37.7 & 48.2 & 26.4 & 38.0 \\
Self-AsK & 23.4 & 32.6 & 15.4 & 22.3 &  5.6 & 12.5 & 39.1 & 51.3 &  41.1 & \underline{52.6} & 28.8 & 40.6 \\
ITER-RETGEN & 25.8 & 36.7 & 16.6 & 23.0 &  9.4 & 16.7 & \underline{46.2} & \underline{58.9} &  39.8 & 51.0 & \underline{\textbf{31.4}} & \underline{43.3} \\
IRCoT & \underline{29.8} & \underline{38.9} & 26.4 & 30.4 &  8.2 & 15.0 & 42.8 & 53.9 &  39.2 & 49.3 & 28.6 & 40.1 \\
\midrule 
CoR & 30.6 & 39.8 & 25.2 & 28.8 &  9.2 & 16.8 & 41.7 & 55.3 &  43.6 & 54.1 & 26.6 & 38.4 \\
ToR & \textbf{38.2} & \textbf{50.4} & \textbf{29.0} & \textbf{37.0} &  13.2 & 22.1 & \textbf{49.2} & \textbf{63.1} &  \textbf{51.0} & \textbf{62.9} & 30.9 & \textbf{43.6} \\
\bottomrule
\end{tabular}
\caption{Answer EM and F1 results on multi-hop question answering datasets. We highlight the best results in bold and underline the best results among other methods.}
\label{tab:answer result}
\end{table*}
\textbf{Direct Prompting} \citep{NEURIPS2020_1457c0d6} prompts a Language Language Model (LLM) to generate the final answer directly. \\
\textbf{CoT prompting} \citep{NEURIPS2022_9d560961}prompts a Language Language Model (LLM) to generate the final answer step by step.\\
\textbf{One-step Retrieval with Direct/ CoT prompting (OneR-Direct/ CoT)} augments Direct/CoT Prompting with paragraphs retrieved by the retriever.\\
\textbf{ReAct/Self-Ask} \citep{yao2023react,press-etal-2023-measuring}  iteratively execute the following steps : (i) Initiate retrieval using the follow-up question generated by the LLM, returning relevant paragraphs, and (ii) Respond to the follow-up question, subsequently deciding whether to generate the next question or finalize the answer. The primary distinction between ReAct and Self-Ask in our implementation lies in the positioning of the retrieved paragraphs.\\
\textbf{ITER-RETGEN} \citep{shao-etal-2023-enhancing} iteratively execute the following steps for several turns:(i) Initiate retrieval using the original question and response generated by the LLM, returning relevant paragraphs, and (ii) Answer the original question with the current turn retrieval paragraphs. Finally, take the last round’s response as an answer. \\
\textbf{IRCoT} \citep{trivedi-etal-2023-interleaving} iteratively execute the following steps: (i) Initiate retrieval using the CoT sentence generated by the LLM, returning relevant paragraphs, and (ii) Generate a new CoT sentence using historical information until a special trigger word is produced or the maximum number of turns is reached. Finally, use the historical retrieval paragraphs to answer the original question.

\subsection{Implementation Details}
\label{chapter:4-4}
We also employed Contriver, GPT-3.5-Turbo, and GPT-4-Turbo for our experiments. We adopted a greedy decoding strategy to ensure the stability of the output. We set the maximum length to 4096 and added as much evidence from the pool without exceeding this limit. We randomly sampled several data from each dataset's training set, manually annotated them for few-shot demonstrations, and adopted a 3-shot setting for all baselines and ours. In the main experiment, the depth of \textsc{ToR} is set to 3, with the number of nodes in each layer being 5, 3, 3. We adopted the evidence-based fusion method, missing paragraph completion expansion strategy, and two kinds of pruning strategies. \\
\textbf{CoR}: To compare the differences between the tree and chain structures, we designed an experiment using the same prompt as \textsc{ToR} but providing only a single reasoning path. The model chooses an action in each iteration and generates new queries based on all retrieved paragraphs. See Table \ref{tab:CoR} for details.

\subsection{Main Result}
As shown by Table \ref{tab:retrieval result} and Table \ref{tab:answer result}, our method \textsc{ToR}  achieves nearly optimal performance in both retrieval metrics and generation metrics across three datasets under two different base models. In the experiments with GPT-4-Turbo as the base model for the three datasets, the retrieval metrics outperform the best-performing baseline method ITER-RETGEN by 13.3\%, 12.0\%, and 1.5\%, respectively. Meanwhile, the F1 values for response generation surpass the highest values among the various baseline methods, with improvements of 9.2\%, 10.3\%, and 0.3\%, respectively.

We consider three reasons for achieving these results: 1)  \textsc{ToR} allows the model to explore multiple reasoning paths, effectively mitigating the cascading errors caused by single reasoning path mistakes. IRCoT, ITER-RETGEN, and CoR (introduced in Section \ref{chapter:4-4}) are all based on chain-of-thought reasoning, and their final response quality is constrained by the accuracy of retrieval and reasoning at each step along the path. In contrast,  \textsc{ToR} employs a tree structure to expand into different paths, sharing the risk of retrieval and reasoning failures. 2) The  \textsc{ToR} structure can effectively reduce the interference of useless information. IRCoT, ITER-RETGEN, and CoR utilize all retrieved paragraphs during the reasoning process, and the useless information contained therein may lead to reasoning errors. We reduce the impact of useless information on retrieval and reasoning along the path by two pruning strategies. 3) \textsc{ToR} enhances the generation quality by improving the quality of retrieval results. Combining the results from Table \ref{tab:retrieval result} and Table \ref{tab:answer result}, we find that retrieval metrics are positively correlated with generation metrics. Therefore, our method improves the final generation quality by enhancing the system's retrieval performance.

Although we adopted the same prompts and experimental settings as in the baseline papers, the results of some baselines on GPT-3.5-Turbo still do not perform well. We speculate that the main reason for this performance gap is the scale of the model parameter. According to the API call prices, GPT-3.5-Turbo costs \$0.5/1M tokens for input and \$1.5/1M tokens for output, and text-davinci-003 costs \$20.0/1M tokens. Based on this, we can infer that the parameter scale of gpt-3.5-turbo is much smaller than that of text-davinci-003.

\begingroup
\setlength{\tabcolsep}{4pt}
\begin{table}[htbp]
  \centering
\begin{tabular}{cccc}
\toprule
Method & HotpotQA & 2WikiMQA & MuSiQue\\
\midrule 
Analysis & 56.8 & 52.0 & 40.4 \\
Paragraph & 64.6 & 62.3 & 46.0 \\
Evidence & 65.5 & 63.7 & 46.2 \\
\bottomrule
\end{tabular}
\caption{ Answer F1 with different evidence fusion strategies. }
\label{tab:evidence fusion}
\end{table}
\endgroup

\textbf{The evidence fusion strategies can enhance the performance of the reader.} As shown by Table \ref{tab:evidence fusion}, generating the final answer based on both retrieved paragraphs and analysis yields optimal performance, demonstrating the effectiveness of our search process.
A significant gap exists between performance derived from analysis and those from paragraphs, indicating that when there are conflicts between different pieces of evidence, the model needs to incorporate information from the retrieved paragraphs to better resolve the contradictions, while the information that analysis can provide is limited.

\begingroup
\setlength{\tabcolsep}{3pt} 
\begin{table}[htbp]
  \centering
\begin{tabular}{ccccccc}
\toprule
\multirow{2}{*}{Method} & \multicolumn{2}{c}{HotpotQA} & \multicolumn{2}{c}{2WikiMQA} & \multicolumn{2}{c}{MuSiQue}\\
\cmidrule(lr){2-3} \cmidrule(lr){4-5} \cmidrule(lr){6-7}
 & Recall & F1 & Recall & F1 & Recall & F1 \\
\midrule 
Direct & 61.5 &57.8 & 58.8& 53.6 & 42.7 & 39.0 \\
CoT & 66.2 & 60.4 & 62.8 & 56.9 & 43.0 & 41.3 \\
MPC & 74.6 & 65.5 & 79.3 & 63.7 & 49.9 & 46.2 \\
\bottomrule
\end{tabular}
\caption{ Paragraphs Recall@15 and Answer F1 with different effective expansion strategies. Direct represents the approach of not using effective expansion and generating a new query directly. CoT represents the approach of using CoT Expansion. MPC represents the approach of using Missing Paragraph Completion Expansion.}
\label{tab: effective expansion}
\end{table}
\endgroup

\begin{table*}[thbp]
  \centering
\begin{tabular}{ccccccccc}
\toprule
Method & \#API  & \#Doc & Rate & \#Evidence &Recall@15 & EM & F1\\
\midrule 
ToR & 16.9 & 15.7 & 92.9 & 2.9 & 79.3 & 51.6 & 63.7\\
w/o repetitive pruning & 33.5 & 18.3& 54.6 & 3.7 & 76.4 & 51.4 & 63.8\\
w/o relevance pruning & 29.1 & 24.6 & 84.5 &3.3 & 73.2 & 48.9 & 59.3 \\
w/o both  & 65.0  & 31.8 & 48.9 & 4.4 & 72.9 & 49.1 & 59.4  \\
\bottomrule
\end{tabular}
\caption{Results of different pruning strategies on 2WikiMQA, \#API represents the average number of GPT API calls. \#Doc represents the average number of different paragraphs retrieved. Rate = \#Doc/\#API, which means the number of reviewed paragraphs per API call, where a higher value indicates more effective API calls (the higher, the better).\#Evidence represents the average number of evidence in the evidence pool. Other metrics are introduced in Section \ref{chapter 4-2}. \textsc{ToR} use both repetitive pruning and relevance pruning. w/o repetitive pruning only uses relevance pruning, and w/o relevance pruning only uses repetitive pruning. w/o, both don't use any pruning strategies.}
\label{tab:pruning}
\end{table*}

\begin{table*}[bthp]
  \centering
\begin{tabular}{cccccccccc}
\toprule
Depth & width & \#API  & Rate & \#Doc & \#Evidence &Recall@15 & EM & F1\\
\midrule 
2 & 5,3 & 10.3 &  10.0 & 97.1 & 1.8 & 69.7 & 44.3 & 55.6\\
3 & 5,3,3 & 16.9 & 15.7 & 92.9 & 2.9 & 79.3 & 51.6 & 63.7\\
4 & 5,3,3,3 & 36.3 & 27.6 & 76.0 & 4.7 & 79.7 & 51.8 & 64.4\\
3 & 10,5,3 & 41.3 & 39.2 & 94.9 & 6.8 & 75.4 & 52.4 & 66.0\\
\bottomrule
\end{tabular}
\caption{Results for different tree depths and widths on 2WikiMQA.}
\label{tab:hyperparameters}
\end{table*}

\textbf{Effective expansion strategies significantly enhance the performance of retrieval.} As shown by Table \ref{tab: effective expansion}, our proposed strategies surpass the baseline strategy, demonstrating their superiority in guiding the search direction by controlling the queries used for retrieval. 
The performance improvement observed with the CoT underscores the significance of incorporating reasoning capabilities into iterative retrieval processes.
Notably, the MPC strategy exhibits the best performance, which may be attributed to the extensive knowledge stored in recent LLMs. This confirms that utilizing both parametric and non-parametric information during the retrieval process can improve retrieval and generation performance \citep{DBLP:conf/iclr/0002IWXJ000023,DBLP:conf/iclr/Sun0TYZ23}.

\textbf{The pruning strategies ensure performance while reducing time cost.} As shown by Table \ref{tab:pruning}, repetitive pruning improves the effective call rate, significantly reducing the time of API calls for the same paragraph and lowering the time cost. Without repetitive pruning, the framework can retrieve more different paragraphs and obtain more evidence through node expansion, which leads to a decrease in the Recall@15 metric. This is because repetitive paragraphs do not provide information gain through node expansion, and the obtained evidence cannot offer additional effective paragraphs, potentially introducing invalid paragraphs that lower retrieval metrics.

Relevance pruning filters out irrelevant paragraphs, reducing ineffective expansion. Without relevance pruning, the framework initiates node expansion for each paragraph. Although this approach can retrieve more different paragraphs, the evidence obtained does not significantly increase, as the retrieval initiated by irrelevant paragraphs does not directly contribute to problem-solving. Additionally, introducing such misleading information may cause the model to generate erroneous reasoning, decreasing Recall@15, EM, and F1 metrics.

\textbf{The depth and width of the tree affect the performance.} As shown by Table \ref{tab:hyperparameters}, we conducted the experiment at different tree depths and widths and drew the following conclusions: 1) As the tree depth increases, our framework retrieves more paragraphs and obtains more evidence, leading to an improvement in retrieval and generation metrics. However, the number of calls also increases non-linearly. This is because our framework generates more feasible paths through node expansion. As this expansion grows exponentially with the increase in tree depth, we need to reasonably limit the depth of the tree to ensure search efficiency. 2) The effective call rate decreases with the deepening of the tree depth. Even though repetitive pruning reduces the repetitive calls of accepted paragraphs, it cannot avoid some unaccepted paragraphs being reviewed multiple times. This phenomenon is amplified with the increase in tree depth. 3) By expanding the breadth of each tree layer, our framework can retrieve more paragraphs and obtain more evidence while ensuring an effective call rate. Notably, its retrieval metrics decrease while its generation metrics improve. We think that the evidence’s proportion of ground truth paragraphs decreases as the breadth increases, leading to fewer recalled ground truth paragraphs at a specific quantity. However, the reader can add more evidence (more than 15) for response generation, thus improving the generation metrics. 4) To balance performance and time cost, we ultimately chose a depth of 3 and widths of 5, 3, and 3.

\section{Conclusion}
This paper proposes \textsc{ToR}, a tree-structured dynamic retrieval framework for multi-hop question-answering tasks. This framework leverages the tree structure and the chain-of-thought capability of Large Language Models(LLMs) to dynamically explore multiple feasible reasoning paths. Experimental results demonstrate that the method effectively explores more diverse reasoning paths while reducing ineffective path expansion. We believe that \textsc{ToR} can serve as a robust baseline model for future research in multi-hop question-answering tasks. Moreover, we hope our framework can be extended to more complex reasoning tasks.

\section*{Limitations}
\textsc{ToR} has requirements for the capabilities of the base models, including 1) The model should have zero-shot or few-shot CoT reasoning abilities. 2) The model should support long-text inputs, as we need to include retrieved paragraphs and few-shot demonstrations in the prompts. 3) The model should have good instruction-following capabilities, as Paragraph Reviews require the model to output intermediate results step-by-step according to the instructions. The model needs to understand the instructions and output in a specific format. Regarding the results returned by LLM, we will parse them according to its prompt. The parsing will fail if the model fails to generate results in that format. The parsing success rate represents LLM's ability to follow complex instructions. Models that meet our requirements tend to have a larger number of parameters.
In contrast, smaller models (with fewer than 20B parameters) often lack satisfactory instruction-following capabilities for our tasks. (with the parsing success rates of output below 85\%, compared to 98.6\% for GPT-3.5-Turbo). This limits the generality of our method. However, as large language models continue to develop, smaller models will meet the above requirements, enhancing our approach’s practicality.

\textsc{ToR} incurs a significant time cost, as our framework calls the LLM at each node, which improves retrieval performance but introduces additional computational overhead. Although we have designed two different pruning strategies to alleviate this issue, an average of 16 LLM calls still exists. In future work, we plan to optimize the framework in the following ways: 1) Implement a more fine-grained repetitive pruning strategy, which involves pruning repetitive paragraphs from multiple perspectives, such as semantic similarity. 2) Develop a more powerful retriever: the experimental results show that reducing tree depth and width can effectively decrease the number of calls, and a more powerful retriever can recall relevant paragraphs more effectively, allowing for a reduction in tree depth and width. 3) Introduce an early termination mechanism: the framework would dynamically choose to terminate the tree search early when the LLM believes sufficient evidence has already been obtained.

Moreover, akin to several other baseline methods with which we have drawn comparisons, our experiments employed the OpenAI LLM API. Owing to the deprecation of the text-davinci-002 API employed by IRCoT and the text-davinci-003 API employed by ITER-RETGEN, we could not employ identical models for a fair comparison. To contrast their approaches, we conducted experiments using the gpt-4-1106-preview and gpt-3.5-turbo-0125 APIs. Although we used the prompts reported in the baseline studies, the issues about prompt transferability precluded a guarantee of fully replicating the effects of their methods. Recognizing that the APIs we have employed may also be deprecated at some point in the future, we intend to release
 all prompts and code to make our research easier to replicate for future study.

Lastly, the performance of \textsc{ToR} on other complex reasoning tasks still requires further verification. We have only validated the effectiveness of the \textsc{ToR} framework on the multi-hop question-answering task. We believe that introducing a tree-like structure in complex reasoning tasks is a viable approach, and we hope that future work can leverage this concept to achieve favorable results in a broader array of complex reasoning tasks.

\section*{Ethical Considerations}
It is well known that Large Language Models(LLMs) suffer from hallucination, privacy, security, and bias during their usage. Although \textsc{ToR} employs retrieval augmentation  that can alleviate the hallucination problem to some extent, it still cannot fully address these issues. Moreover, our framework does not consider bias, security, and privacy concerns. If our framework is to be deployed in practical application scenarios, certain restrictions should be implemented to prevent generating harmful information.

\newpage

\bibliography{custom}
\newpage
\appendix

\section{Few-Shot Prompts}

This section presents all the few-shot prompts used in our experiments.
\subsection{ToR}
Our ToR framework involves the following prompts: Paragraphs Review (with CoT expansion) prompt in Table \ref{tab:PR-CoT}; Paragraphs Review (without effective expansion) prompt in Table \ref{tab:PR-Direct}; Missing Paragraph Completion expansion prompt in Table \ref{tab:MPC}; three evidence fusion prompts in Table \ref{tab:A-based-EF}, \ref{tab:P-based-EF}, \ref{tab:E-based-EF} and CoR prompt in Table \ref{tab:CoR}.

\begin{table*}[thbp]
  \centering
\begin{tabular}{p{16cm}}
\toprule
\textbf{Instruction} \\
I will give you a query and few of documents. Your ultimate task is to judge if the documents support answering the question. However, you need to think step by step. Follow these steps to answer question. The most important thing you must aware, these documents are a whole. So in later steps, if you are asked to judge something, please judge with all the documents. \\
Step 1 \\
You need to judge if the documents are relevant to the question. Please pay attention that this is a multi-hop QA task, so if these document contain any useful information which can help you get closer to the final answer, you have to output [RELEVANT], only if those documents don't contain any useful information you have to output [IRRELEVANT]. \\
Output: \\
- Thought: why you think these documents are relevant to question or not. \\
- Judgment: Please pay attention that this is a multi-hop QA task, so if these document contain any useful information which can help you get closer to the final answer, you have to output [RELEVANT], only if those documents don't contain any useful information you have to output [IRRELEVANT]. [RELEVANT] if the documents are relevant to the question, otherwise [IRRELEVANT]. \\
Output Format: \\
- Thought: A few words for judgment. \\
- Judgment: [RELEVANT] or [IRRELEVANT] \\
Step 2 \\
You need to judge if the documents contain enough information to answer the question. Please pay attention that this is a multi-hop QA task, so please double-check that you have all the information you need to answer the question. Whenever you find that you miss any information to answer the question you need to output [UNSUPPORTED]. Only if you get all the information supported to answer the question, you have to output [SUPPORTED].\\
Output: 
- Thought: If you are confident that you have all the information to answer the question, briefly write the reasoning path. Otherwise, list what kind of information is important to answer this question. \\
- Judgment: [SUPPORTED] if contain enough information, otherwise [UNSUPPORTED].\\
Output Format: 
- Thought: A few words for judgment. \\
- Judgment: [SUPPORTED] or [UNSUPPORTED], \\
Step 3\\
If you output [SUPPORTED] in step2 ,you need to answer the question with these documents, otherwise you need to think what kind of extra information you need to answer the question and output new query. 
Output: \\
- Thought: if you can answer the question according to step2. If you can't, think what extra information you need. \\
- Output: [ANSWER] if you can answer the question, otherwise [QUERY] \\
Output Format: \\
- Thought: A few words for thought. \\
- Output: A special token ([ANSWER] or [QUERY]) follow with your answer or new query. \\
\midrule 
\textbf{Prompt Format} \\
Instruction:\{INST\} \\
Demonstration:\{DEMO\} \\
Question:\{Q\} \\
Documents: \{D\} \\

\bottomrule
\end{tabular}
\caption{Paragraphs Review (with CoT expansion) prompt. \{INST\} will be replaced by Instruction, \{DEMO\} will be replaced by 3-shot demonstrations, \{Q\} will be replaced by question and \{D\} will be replaced by the paragraphs on the reasoning path.}
\label{tab:PR-CoT}
\end{table*}

\begin{table*}[thbp]
  \centering
\begin{tabular}{p{16cm}}
\toprule
\textbf{Instruction Format} \\
I will give you a query and few of documents. Your ultimate task is to judge if the documents support answering the question. However, you need to think step by step. Follow these steps to answer question. The most important thing you must aware, these documents are a whole. So in later steps, if you are asked to judge something, please judge with all the documents. \\
Step 1 \\
You need to judge if the documents are relevant to the question. Please pay attention that this is a multi-hop QA task, so if these document contain any useful information which can help you get closer to the final answer, you have to output [RELEVANT], only if those documents don't contain any useful information you have to output [IRRELEVANT]. \\
Output: \\
- Judgment: Please pay attention that this is a multi-hop QA task, so if these document contain any useful information which can help you get closer to the final answer, you have to output [RELEVANT], only if those documents don't contain any useful information you have to output [IRRELEVANT]. [RELEVANT] if the documents are relevant to the question, otherwise [IRRELEVANT]. \\
Output Format: \\
- Judgment: [RELEVANT] or [IRRELEVANT] \\
Step 2 \\
You need to judge if the documents contain enough information to answer the question. Please pay attention that this is a multi-hop QA task, so please double-check that you have all the information you need to answer the question. Whenever you find that you miss any information to answer the question you need to output [UNSUPPORTED]. Only if you get all the information supported to answer the question, you have to output [SUPPORTED].\\
Output: 
- Judgment: [SUPPORTED] if contain enough information, otherwise [UNSUPPORTED].\\
Output Format: 
- Judgment: [SUPPORTED] or [UNSUPPORTED], \\
Step 3\\
If you output [SUPPORTED] in step2 ,you need to answer the question with these documents, otherwise you need to think what kind of extra information you need to answer the question and output new query. 
Output: \\
- Output: [ANSWER] if you can answer the question, otherwise [QUERY] \\
Output Format: \\
- Output: A special token ([ANSWER] or [QUERY]) follow with your answer or new query. \\
\midrule 
\textbf{Prompt Format} \\
Instruction:\{INST\} \\
Demonstration:\{DEMO\} \\
Question:\{Q\} \\
Documents: \{D\} \\

\bottomrule
\end{tabular}
\caption{Paragraphs Review (without effective expansion) prompt.\{INST\} will be replaced by Instruction, \{DEMO\} will be replaced by 3-shot demonstrations, \{Q\} will be replaced by question and \{D\} will be replaced by the paragraphs on the reasoning path.}
\label{tab:PR-Direct}
\end{table*}

\begin{table*}[thbp]
  \centering
\begin{tabular}{p{16cm}}
\toprule
\textbf{Instruction} \\
 I will give you a question and few of references, your ultimate task is answer the question according to these references. However these references lack of some important information to answer the question. So you should follow these two steps to answer the question. \\
 Step 1 \\
 Complete the reference information In this step you need to generate the missing information based on your knowledge so that you can answer the question correctly. \\
 Output: \\
 - Thought: To be effective generate missing information, you must analyze what information is currently missing, based on the question and references. \\
 - Information: A special token ([INFO]) follow with the missing information you want to supply. \\
 Step 2 \\
 Answer generation You need to answer the question based on references I gave you and the information you supplied by yourself. \\
 Output: \\
 - Answer:A special token ([ANSWER]) follow with the answer to my question.\\
\midrule 
\textbf{Prompt Format} \\
Instruction:\{INST\} \\
Demonstration:\{DEMO\} \\
Question:\{Q\} \\
References: \{R\} \\

\bottomrule
\end{tabular}
\caption{Missing Paragraph Completion expansion prompt.\{INST\} will be replaced by Instruction, \{DEMO\} will be replaced by 3-shot demonstrations, \{Q\} will be replaced by question and \{R\} will be replaced by the paragraphs on the reasoning path.}
\label{tab:MPC}
\end{table*}

\begin{table*}[thbp]
  \centering
\begin{tabular}{p{16cm}}
\toprule
\textbf{Instruction} \\
 Your task is to answer my questions in a few words based on the relevant documents I have provided .If documents not provide enough information to answer the question, answer it by yourself. Your response can contain several sentences, but The last sentence must include "The answer is xxx. \\
\midrule 
\textbf{Prompt Format} \\
Instruct:\{INST\} \\
Demonstration:\{DEMO\} \\
Documents:\{D\} \\
Question:\{Q\} \\

\bottomrule
\end{tabular}
\caption{Paragraph-based Fusion prompt.\{INST\} will be replaced by Instruction, \{DEMO\} will be replaced by 3-shot demonstrations, \{D\} will be replaced by the paragraphs in evidence pool and \{Q\} will be replaced by question.}
\label{tab:P-based-EF}
\end{table*}

\begin{table*}[thbp]
  \centering
\begin{tabular}{p{16cm}}
\toprule
\textbf{Instruction} \\
 Your task is to answer my questions based on some assertions that I provide related to the problem. There may be some conflicts among these assertions, so use all the assertions and your own knowledge to make a judgment. Your response can contain several sentences, but The last sentence must include "The answer is xxx. \\
\midrule 
\textbf{Prompt Format} \\
Instruct:\{INST\} \\
Demonstration:\{DEMO\} \\
Assertions:\{A\} \\
Question:\{Q\} \\

\bottomrule
\end{tabular}
\caption{Analysis-based Fusion prompt.\{INST\} will be replaced by Instruction, \{DEMO\} will be replaced by 3-shot demonstrations, \{A\} will be replaced by the short analysis in evidence pool and \{Q\} will be replaced by question.}
\label{tab:A-based-EF}
\end{table*}

\begin{table*}[thbp]
  \centering
\begin{tabular}{p{16cm}}
\toprule
\textbf{Instruction} \\
 Your task is to answer my questions based on a set of evidence that I provide. Each piece of evidence includes an assertion related to the question and several reference documents supporting the assertion. There may be some conflicts among the assertions in the evidence, so use all the evidence, supporting reference documents, and your own knowledge to make a judgment. Your response can contain several sentences, but The last sentence must include "The answer is xxx. \\
\midrule 
\textbf{Evidence Format} \\
Assertions:\{A\} \\
Documents:\{D\} \\
\midrule 
\textbf{Prompt Format} \\
Instruct:\{INST\} \\
Demonstration:\{DEMO\} \\
Evidence:\{E\} \\
Question:\{Q\} \\

\bottomrule
\end{tabular}
\caption{Evidence-based Fusion prompt. The evidence in the evidence pool will be organized according to the format shown in the \textbf{Evidence Format}.\{INST\} will be replaced by Instruction, \{DEMO\} will be replaced by 3-shot demonstrations, \{E\} will be replaced by evidence, and \{Q\} will be replaced by question.}
\label{tab:E-based-EF}
\end{table*}

\begin{table*}[thbp]
  \centering
\begin{tabular}{p{16cm}}
\toprule
\textbf{Instruction} \\
I will give you a query and few of documents. Your ultimate task is to judge if these documents support answering the question. However, you need to think step by step. Follow these steps to answer question. \\
Step 1 \\
You need to judge if the documents are relevant to the question. Please pay attention that this is a multi-hop QA task, so if these document contain any useful information which can help you get closer to the final answer, you have to output [RELEVANT], only if those documents don't contain any useful information you have to output [IRRELEVANT]. \\
Output: \\
- Thought: why you think these documents are relevant to question or not. \\
- Judgment: Please pay attention that this is a multi-hop QA task, so if these document contain any useful information which can help you get closer to the final answer, you have to output [RELEVANT], only if those documents don't contain any useful information you have to output [IRRELEVANT]. [RELEVANT] if the documents are relevant to the question, otherwise [IRRELEVANT]. \\
Output Format: \\
- Thought: A few words for judgment. \\
- Judgment: [RELEVANT] or [IRRELEVANT] \\
Step 2 \\
You need to judge if the documents contain enough information to answer the question. Please pay attention that this is a multi-hop QA task, so please double-check that you have all the information you need to answer the question. Whenever you find that you miss any information to answer the question you need to output [UNSUPPORTED]. Only if you get all the information supported to answer the question, you have to output [SUPPORTED].\\
Output: 
- Thought: If you are confident that you have all the information to answer the question, briefly write the reasoning path. Otherwise, list what kind of information is important to answer this question. \\
- Judgment: [SUPPORTED] if contain enough information, otherwise [UNSUPPORTED].\\
Output Format: 
- Thought: A few words for judgment. \\
- Judgment: [SUPPORTED] or [UNSUPPORTED], \\
Step 3\\
If you output [SUPPORTED] in step2 ,you need to answer the question with these documents, otherwise you need to think what kind of extra information you need to answer the question and output new query. 
Output: \\
- Thought: if you can answer the question according to step2. If you can't, think what extra information you need. \\
- Output: [ANSWER] if you can answer the question, otherwise [QUERY] \\
Output Format: \\
- Thought: A few words for thought. \\
- Output: A special token ([ANSWER] or [QUERY]) follow with your answer or new query. \\
\midrule 
\textbf{Prompt Format} \\
Instruction:\{INST\} \\
Demonstration:\{DEMO\} \\
Question:\{Q\} \\
Documents: \{D\} \\

\bottomrule
\end{tabular}
\caption{CoR prompt. Attention that \{D\} will be replaced by all the retrieved paragraphs.}
\label{tab:CoR}
\end{table*}

\end{document}